# Lightweight Retrieval-Augmented Generation and Large Language Model-Based Modeling for Scalable Patient–Trial Matching

Xiaodi Li, Ph.D.[1,*], Yang Xiao[2,*], Munhwan Lee, Ph.D.[1], Konstantinos Leventakos, M.D., Ph.D.[3], Young J. Juhn, M.D., M.P.H.[4], David Jones, M.D.[5], Terence T. Sio, M.D., M.S.[6], Wei Liu, Ph.D.[3], Maria Vassilak, M.D., Ph.D.[7], and Nansu Zong, Ph.D. [1,+]

[1]Department of Artificial Intelligence and Informatics, Mayo Clinic, Rochester, MN, USA; [2]Computer Science Department, University of Tulsa, Tulsa, OK, USA; [3]Mayo Clinic Comprehensive Cancer Center, Mayo Clinic, Rochester, MN, USA; [4]Division of Community Pediatric and Adolescent Medicine, Department of Pediatrics, Mayo Clinic, Rochester, MN, USA; [5]Department of Neurology, Mayo Clinic, Rochester, MN, USA; [6]Department of Radiation Oncology, Mayo Clinic, Rochester, MN, USA; [7]Department of Quantitative Health Sciences, Mayo Clinic, Rochester, MN, USA

* Equal Contribution

+ Corresponding author. Email: Zong.Nansu@mayo.edu

## Abstract

Patient–trial matching requires reasoning over long, heterogeneous electronic health records (EHRs) and complex eligibility criteria, posing significant challenges for scalability, generalization, and computational efficiency. Existing approaches either rely on full-document processing with large language models (LLMs), which is computationally expensive, or use traditional machine learning methods that struggle to capture unstructured clinical narratives. In this work, we propose a lightweight framework that combines retrieval-augmented generation and large language model-based modeling for scalable patient–trial matching. The framework explicitly separates two key components: retrieval-augmented generation is used to identify clinically relevant segments from long EHRs, reducing input complexity, while large language models are used to encode these selected segments into informative representations. These representations are further refined through dimensionality reduction and modeled using lightweight predictors, enabling efficient and scalable downstream classification. We evaluate the proposed approach on multiple public benchmarks (n2c2, SIGIR, TREC 2021/2022) and a real-world multimodal dataset from Mayo Clinic (MCPMD). Results show that retrieval-based information selection significantly reduces computational burden while preserving clinically meaningful signals. We further demonstrate that frozen LLMs provide strong representations for structured clinical data, whereas fine-tuning is essential for modeling unstructured clinical narratives. Importantly, the proposed lightweight pipeline achieves performance comparable to end-to-end LLM approaches with substantially lower computational cost. Overall, this work provides a practical and privacy-preserving solution for deploying LLM-based systems in real-world clinical trial recruitment, highlighting the importance of efficient information selection and representation learning for scalable healthcare applications.

## Author Summary

Matching patients to clinical trials is essential for developing new treatments, but it remains a slow and challenging process. In practice, clinicians and research staff must review large amounts of patient information and compare it with complex trial requirements, which is time-consuming and prone to error.

A key challenge is that patient records are often long and contain both structured data and free-text clinical notes, making it difficult to efficiently identify relevant information. In this study, we developed a lightweight and scalable artificial intelligence approach that focuses on selecting and analyzing the most relevant parts of patient records rather than processing entire documents. Our method integrates retrieval techniques with large language models, enabling efficient identification of clinically meaningful information while reducing unnecessary computation. By combining targeted information selection, compact representations, and simple prediction models, the approach achieves accurate patient–trial matching with lower computational cost. We evaluated this method on multiple datasets, including real-world clinical data, and found that it performs competitively while requiring fewer resources than existing methods. Overall, our work provides a practical and scalable solution to improve clinical trial recruitment.

## 1. Introduction

Matching patients to appropriate clinical trials is a critical yet challenging task in clinical research, and more fundamentally reflects the problem of efficiently modeling long, heterogeneous electronic health records (EHRs) under complex eligibility constraints. While trials are essential for evaluating new therapies and advancing medical science, patient recruitment remains a major obstacle, frequently resulting in delays or even failures[1]. Traditional recruitment methods based on manual screening require staff to review EHRs against complex eligibility criteria, highlighting the difficulty of extracting clinically relevant information from long and unstructured patient records, which is highly time-consuming (taking up to 1 hour per patient[1]) and prone to substantial subjective errors[2, 3]. Automated screening methods based on intelligent technologies have been proposed to address these challenges.

Recently, machine learning (ML) has demonstrated strong potential in natural language processing (NLP), including applications in patient–trial matching, by enabling automated modeling of clinical text and structured EHR data[4, 5]. This is because the vast volume of unstructured clinical text within EHRs, including physician notes, diagnostic reports, and medication histories[6], for example, each clinical trial specifies a detailed set of eligibility criteria that all enrolled patients must fully satisfy[7]. Thus, the ML models can automatically process the EHRs data. Several approaches have been proposed to automate patient–trial matching. For example, ELIXR integrated multiple models into one workflow including visual task and textual task[8]. EliIE extract relation from 230 Alzheimer's clinical trials via Support Vector Machine (SVM)[9]. Criteria2Query directly transforms free-text eligibility criteria into executable cohort queries under the OMOP Common Data Model[10].

Although they were able to achieve relatively impressive results on their respective datasets and tasks, these traditional machine learning or natural language processing methods they employed often required extensive maintenance and struggled with adaptability[11], such as some models fail to generalize to entirely new data beyond the training set due to their idiosyncratic grammar and terminology[12]. Also, there is another primary reason behind this limitation is that many of the existing methods were originally designed to handle structured data, while in practice, the majority of EHRs are stored in unstructured form such as free-text clinical notes, diagnostic reports, and medication histories[3, 11]. This mismatch between data type and methodological design significantly restricts effectiveness, as these approaches do not explicitly address how to select and represent clinically meaningful information from long and noisy EHR sequences.

Building upon this direction, the rapid development of large language models (LLMs) has further opened up new opportunities for modeling unstructured clinical narratives and complex eligibility criteria[3]. At the same time, retrieval-augmented generation (RAG) has emerged as a key mechanism for improving the efficiency of LLM-based approaches by selectively incorporating clinically relevant information rather than processing full EHR records. Recent studies have explored LLM-based patient–trial matching, such as TrialGPT, which achieved high accuracy but lacked a zero-shot evaluation component[2], while[1] directly deployed commercial models on the open-source n2c2 dataset, but lacks validation on real-world data.

Despite these advances, current RAG-augmented LLM approaches face several practical limitations related to scalability, generalization, and computational efficiency. First, although LLMs support extended context lengths, practical constraints related to input formatting, computational efficiency, and information density limit the effective utilization of comprehensive clinical data, making full-document processing inefficient and difficult to scale. In real-world EHRs, a single patient record often spans thousands of tokens across multiple notes, making it infeasible to input the complete longitudinal history without truncation or aggressive summarization. Second, many existing methods are evaluated primarily on curated or task-specific datasets, and their performance does not necessarily translate to heterogeneous, noisy real-world EHR data, highlighting challenges in model adaptation. Third, end-to-end processing of long clinical documents with large models incurs substantial computational cost, limiting scalability for large patient cohorts and real-world deployment. In addition, existing RAG-based approaches often rely on synthetic or open-source datasets (e.g., n2c2), leaving their effectiveness on real-world EHR data insufficiently validated, and approaches built on commercial LLMs (GPT-4[13], Meta Llama[14]) raise concerns regarding data privacy, security, and limited control over model behavior and data governance.

These limitations highlight the need for a lightweight framework that jointly integrates retrieval-based information selection with LLM-based representation learning, enabling efficient, scalable, and adaptable patient–trial matching in real-world clinical settings.

To address these challenges, we propose a lightweight retrieval-augmented generation and large language model-based framework for scalable patient–trial matching. The proposed approach is designed around a lightweight principle, where retrieval-based information selection, compact representation learning, and efficient prediction are jointly optimized. By selectively incorporating clinically relevant information, the framework avoids full-document processing and enables efficient modeling of long, multimodal patient records. Retrieved EHR segments are encoded using lightweight, locally deployable LLMs to generate embeddings, which are further refined through dimensionality reduction to produce compact representations. These representations are then processed by lightweight downstream predictors, decoupling representation learning from prediction and significantly reducing computational cost. Importantly, this lightweight design not only improves scalability, but also enhances adaptability to heterogeneous and noisy real-world EHR data, as retrieval reduces irrelevant context and stabilizes representation learning. In addition, the framework supports efficient large-scale and privacy-preserving deployment by avoiding reliance on large proprietary models.

This study makes three contributions to lightweight, scalable, and efficient patient–trial matching. (1) We propose a lightweight retrieval-augmented framework that enables scalable modeling of long, multimodal EHRs by selecting task-relevant clinical text and combining it with structured data, reducing input complexity and making it feasible to apply LLM-based models to real-world longitudinal records. (2) Through extensive experiments on both public benchmarks and a real-world Mayo Clinic dataset (MCPMD), we systematically demonstrate how lightweight LLM-based representations behave across modalities, showing that frozen LLMs are sufficient for structured clinical variables, whereas fine-tuning is critical for extracting predictive signal from unstructured clinical notes, especially in mixed-modality settings. (3) We show that a fully lightweight pipeline, integrating retrieval-augmented LLM representations with dimensionality reduction and efficient downstream classifiers, achieves competitive or superior performance with substantially lower computational cost than end-to-end fine-tuning, while improving robustness and highlighting the challenges of generalization across heterogeneous trials.

## 2. Results

### 2.1. Settings

*1) <u>Datasets</u>*

We use the open-source datasets n2c2[15], the Special Interest Group on Information Retrieval (SIGIR) 2016[16], TREC 2021[17], and TREC 2022[18], and private dataset MCPMD from Mayo Clinic.

The 2018 n2c2 Clinical Trial Cohort Selection dataset was developed for automated eligibility screening research and includes EHR data from 288 patients with diabetes (202 training, 86 test). Each patient has 2–5 de-identified clinical notes with an average length of 2,711 words, collected from Mass General and Brigham and Women's hospitals. The dataset covers 13 predefined inclusion criteria commonly used in trial eligibility assessment, with labels annotated by two medical experts as "MET" or "NOT MET" (Cohen's $\kappa = 0.54$). This dataset represents a synthetic trial scenario rather than real-world clinical enrollment.

The SIGIR 2016 dataset defines three classes, irrelevant, potential, and eligible, reflecting increasing likelihood of trial referral. The TREC 2021 and 2022 datasets similarly use three categories: irrelevant, excluded/ineligible, and eligible. For binary classification, we merged the potential and eligible categories into a single positive class.

The MCPMD contains both structured and unstructured data collected from multiple clinical trials, enabling comprehensive patient characterization. For our experiments, we randomly selected five trials (NCT01767909, NCT02008357, NCT02565511, NCT02669433, and NCT04468659). The structured data consist of relational tabular records, including patient demographics, diagnoses, medications, allergies, flowsheets, and radiology information, with associated identifiers and timestamps that support systematic analysis. The unstructured data comprise free-text clinical notes authored by healthcare providers, capturing clinical reasoning and contextual details not fully represented in structured fields. Together, these modalities provide a rich foundation for patient–trial matching that integrates rule-based filtering with natural language understanding.

Trial NCT01767909 examines the effects of intranasally-administered insulin on cognition, entorhinal cortex and hippocampal atrophy, and cerebrospinal fluid (CSF) biomarkers in amnestic mild cognitive impairment (aMCI) or mild Alzheimer's disease (AD). Trial NCT02008357 tests whether an investigational drug called solanezumab can slow the progression of memory problems associated with brain amyloid (protein that forms plaques in the brains of people with AD). Trial NCT02565511 tests whether two investigational drugs called CAD106 and CNP520, administered separately, could slow down the onset and progression of clinical symptoms associated with AD in participants at the risk to develop clinical symptoms based on their age and genotype. Trial NCT02669433 evaluates the efficacy and safety of intepirdine (RVT-101) in patients with dementia with Lewy bodies. Trial NCT04468659 tries to determine whether treatment with lecanemab is superior to placebo on change from baseline of the Preclinical Alzheimer Cognitive Composite 5 (PACC5) at 216 weeks of treatment (A45 Trial) and to determine whether treatment with lecanemab is superior to placebo in reducing brain amyloid accumulation as measured by amyloid positron emission tomography (PET) at 216 weeks of treatment (A3 Trial). This study also evaluates the long-term safety and tolerability of lecanemab in participants enrolled in the Extension Phase.

The raw data of the private dataset consist of both structured and unstructured components. We applied three types of processing: (1) processing structured raw data alone, (2) processing unstructured raw data alone, and (3) combining both types of data for joint processing.

Table 1 summarizes the availability of eligibility criteria across five clinical trials when mapped to EHR data. Demographic information is consistently well captured, with all trials achieving complete availability (100% Pos and Idf), indicating that basic patient characteristics can be reliably extracted using structured SQL queries, while other structured clinical concepts, such as diagnoses, medications, and hospitalization records, are often identifiable through rule-based approaches or string matching on standardized codes and descriptions. In contrast, several inclusion criteria, such as cognitive assessments, gene testing, and family history, are largely unobservable with near-zero availability, reflecting limitations of routine clinical documentation. Imaging-based biomarkers and medication-related inclusion criteria show partial availability in specific trials, highlighting heterogeneity in data capture. For exclusion criteria, diagnosis-related information demonstrates relatively high availability across all trials, while other domains exhibit substantial variability: medication and hospitalization criteria are often identifiable (high Idf) but not always associated with a high proportion of qualifying patients (lower Pos), and categories such as surgical history and allergy remain largely unobserved. Psychiatric/psychological conditions and life habit factors show moderate but inconsistent availability, underscoring gaps in structured recording of behavioral and contextual health information. Overall, Table 1 illustrates that while many criteria can be retrieved using structured queries or string matching, a substantial portion of trial-specific eligibility criteria remains partially or entirely unobservable; therefore, for these non-searchable components, we leverage embedding-based patient matching to capture latent clinical information from unstructured data and improve alignment with trial requirements.

**Table 1.** Availability of Eligibility Criteria in Five Clinical Trials. White cells indicate criteria that are not searchable in EHR data. Pos = proportion of patients identified with a valid value for the criterion; Idf = proportion of patients with available information from the data source.

| Drugs | Insulin | | Solanezumab | | CAD106, CNP520 | | RVT-101 | | Lecanemab | |
|---|---|---|---|---|---|---|---|---|---|---|
| NCT number | NCT01767909 | | NCT02008357 | | NCT02565511 | | NCT02669433 | | NCT04468659 | |
| Screened patients | N=30 | | N=114 | | N=39 | | N=23 | | N=158 | |
| Inclusion Criteria: | Pos/Idf | Idf/N | Pos/Idf | Idf/N | Pos/Idf | Idf/N | Pos/Idf | Idf/N | Pos/Idf | Idf/N |
| Demographic | 100.00 % | 100.00 % | 100.00 % | 100.00 % | 100.00 % | 100.00 % | 100.00 % | 100.00 % | 100.00 % | 100.00 % |
| Cognitive test | 0.00% | 0.00% | 0.00% | 0.00% | 0.00% | 0.00% | 0.00% | 0.00% | 0.00% | 0.00% |
| MRI/PET marker | 0.00% | 0.00% | 0.00% | 36.00 % | 0.00% | 0.00% | 0.00% | 0.00% | 0.00% | 0.00% |
| Family history | 0.00% | 0.00% | 0.00% | 0.00% | 0.00% | 0.00% | 0.00% | 0.00% | 55.00 % | 38.00 % |
| Gene test | 0.00% | 0.00% | 0.00% | 0.00% | 0.00% | 0.00% | 0.00% | 0.00% | 0.00% | 0.00% |
| Medication | 0.00% | 0.00% | 0.00% | 0.00% | 0.00% | 0.00% | 66.70 % | 91.30 % | 0.00% | 0.00% |
| Exclusion Criteria: | | | | | | | | | | |
| Diagnosis | 53.30 % | 100.00 % | 37.30 % | 96.50 % | 26.90 % | 66.70 % | 52.60 % | 82.60 % | 37.90 % | 91.80 % |
| Medication | 40.00% | 100.00% | 0.90% | 98.20% | 0.00% | 100.00% | 0.00% | 0.00% | 0.00% | 0.00% |
| Surgical | 0.00% | 0.00% | 0.00% | 0.00% | 0.00% | 0.00% | 0.00% | 0.00% | 0.00% | 0.00% |
| Hospitalization | 0.00% | 100.00% | 0.90% | 100.00% | 0.00% | 100.00% | 0.00% | 100.00% | 2.50% | 100.00% |
| Psychiatric/psychological | 13.30% | 100.00% | 5.50% | 96.50% | 0.00% | 0.00% | 0.00% | 0.00% | 9.00% | 91.80% |
| Life habit | 0.00% | 0.00% | 0.00% | 96.50% | 0.00% | 66.70% | 0.00% | 0.00% | 0.70% | 91.80% |
| Allergy/Hypersensitivity | 0.00% | 0.00% | 0.00% | 0.00% | 0.00% | 0.00% | 0.00% | 0.00% | 0.00% | 53.80% |

### 2) *Evaluation Metrics*

We evaluate our models based on multiple performance metrics to comprehensively assess their effectiveness inpatient-trial matching. Our primary evaluation focuses on precision, recall, and Macro-F1 scores for the binary classification task of determining whether a patient meets each of the eligibility criteria. These metrics provide insights into the model's ability to accurately classify patients while balancing both sensitivity and specificity.

## 2.2. Experimental Results

We evaluate the proposed framework across six tasks designed to assess the impact of representation strategy, LLM backbone, dimensionality reduction, fine-tuning, and generalization across datasets and trials. Experiments are conducted on the MCPMD and multiple open-source benchmarks, using structured, unstructured, and mixed EHR inputs.

### 1) *Task 1: Effect of Representation and Classification Strategy*

We first examine how different downstream processing strategies perform when applied to the same EHR representation across structured-only, unstructured-only, and mixed data settings. As shown in Figure 1, substantial performance differences arise solely from variations in how the shared representation is processed and classified.

Pipelines that directly feed high-dimensional representations into standard classifiers exhibit less stable performance, particularly for unstructured and mixed inputs. In contrast, pipelines that incorporate dimensionality reduction followed by a lightweight MLP classifier consistently achieve higher performance across all evaluation metrics. The improvements are most pronounced in unstructured and mixed settings. These results demonstrate that, even with identical representations, downstream processing design plays a critical role in effective patient–trial matching.

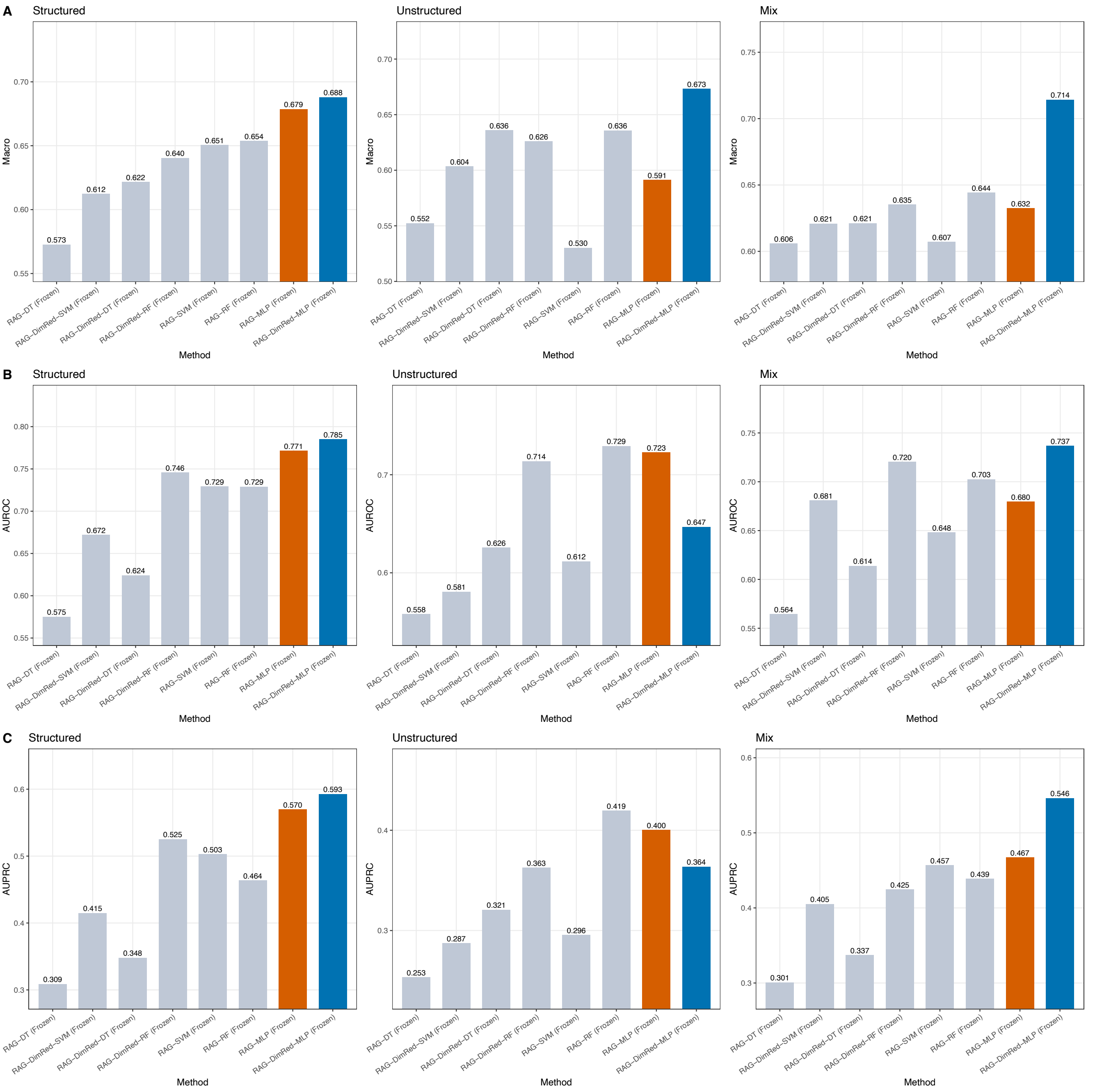

**Figure 1.** Task 1 results on MCPMD: Comparison of downstream processing strategies applied to a shared EHR representation across data modalities. This figure compares patient–trial matching performance when applying different downstream processing strategies to the same underlying EHR representation. Columns correspond to structured, unstructured, and mixed EHR inputs, reflecting increasing data complexity. Rows (A–C) report performance measured by AUROC, AUPRC, and Macro-F1, respectively. Each bar represents a complete pipeline that shares the same representation construction step but differs in downstream components, including the use of dimensionality reduction (DimRed) and the choice of classifier. This comparison isolates the effect of downstream processing choices on performance across EHR modalities.

### 2) *Task 2: Effect of LLM Backbone Choice*

Next, we evaluate the impact of different LLM backbones under a fixed RAG-DimRed-MLP pipeline, where all components other than the LLM backbone are held constant. Figure 2 summarizes performance across structured-only, unstructured-only, and mixed EHR settings using Macro-F1, AUROC, and AUPRC. Across all data modalities, both Mistral-7B and Falcon-7B achieve competitive performance, indicating that both backbones are capable of supporting the patient–trial matching task within the same pipeline design. However, their performance profiles differ across metrics and input types. Falcon-7B attains slightly higher AUROC and AUPRC in several structured and unstructured settings, but its performance varies more substantially across modalities and metrics. In contrast, Mistral-7B exhibits more consistent performance across all three data settings and achieves higher Macro-F1 in the mixed setting, which places greater emphasis on balanced classification under class imbalance and most closely reflects real-world patient–trial matching scenarios. Based on its overall stability and stronger performance in the mixed-data setting, we select Mistral-7B as the primary backbone for subsequent experiments, while retaining Falcon-7B as a comparative baseline.

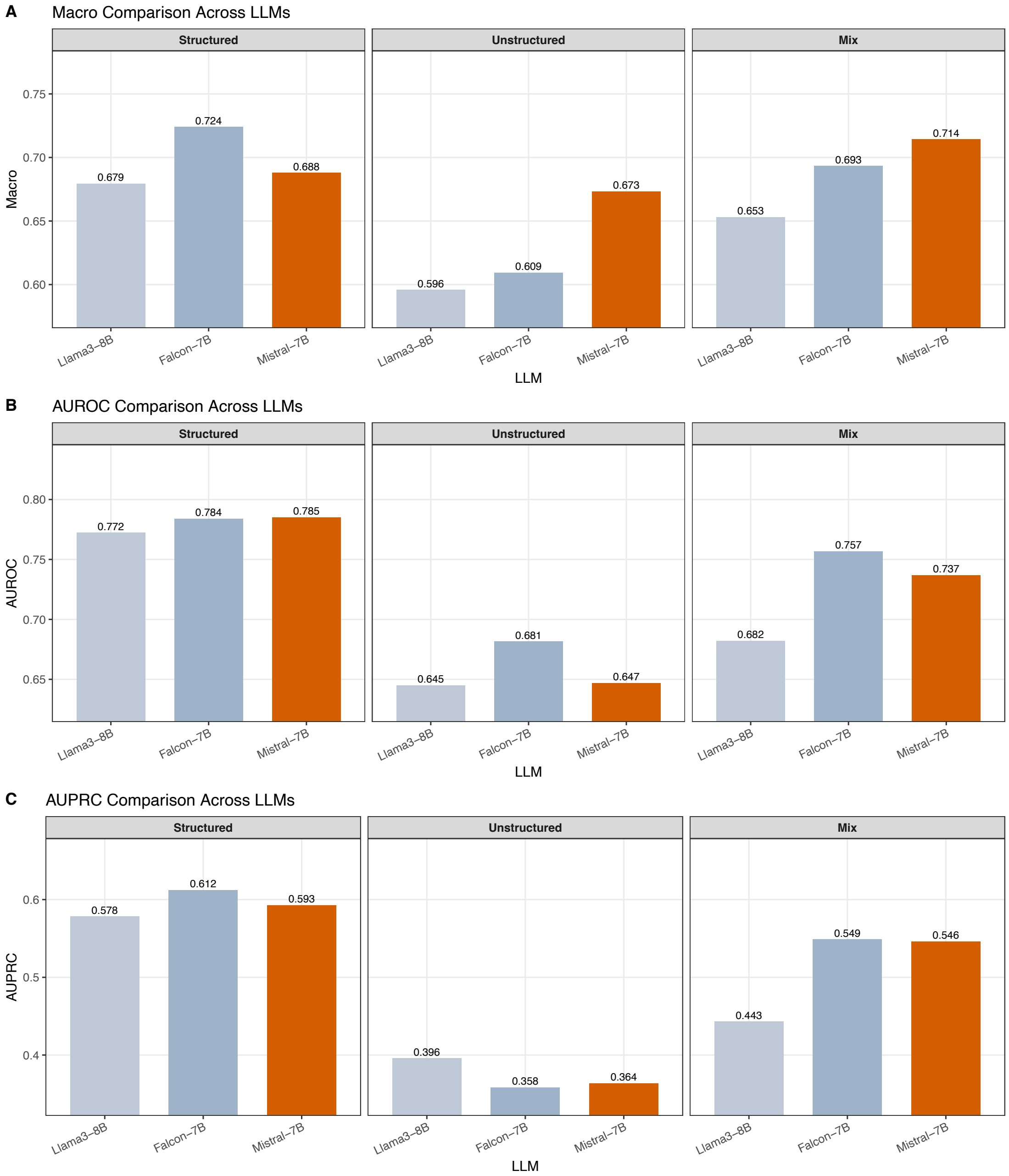

**Figure 2.** Task 2 results on MCPMD: Comparison of different LLM backbones within the RAG-DimRed-MLP pipeline across EHR modalities. This figure compares the performance of different LLM backbones when integrated into the RAG-DimRed-MLP patient–trial matching pipeline on the MCPMD dataset. Columns correspond to structured, unstructured, and mixed EHR inputs. Rows (A–C) report results measured by Macro-F1, AUROC, and AUPRC, respectively. Across all settings, the retrieval strategy, dimensionality reduction module, and downstream MLP classifier are held fixed, with only the LLM backbone varied. This comparison isolates the effect of LLM choice on downstream matching performance under a consistent pipeline design.

*3) Task 3: Effect of Dimensionality Reduction Strategy*

We next examine how different dimensionality reduction (DimRed) strategies affect downstream patient–trial matching performance when applied to the same LLM representations. As shown in Figure 3, we evaluate multiple representation variants, including DimRed applied along the sequence-length dimension, DimRed applied along the hidden-state dimension with varying numbers of components, last-token embeddings, and hybrid approaches that combine DimRed and last-token representations. Clear performance differences are observed across these strategies. Methods that rely solely on last-token embeddings yield lower and less stable performance across all evaluation metrics, indicating that a single-token summary is insufficient to capture the full clinical signal. DimRed applied along the hidden dimension improves performance as the number of retained components increases, with the strongest results achieved at 128 components, while more aggressive compression leads to performance degradation. In contrast, DimRed applied along the sequence-length dimension with a single component consistently achieves strong performance across Macro-F1, AUROC, and AUPRC. This behavior suggests that reducing along the sequence dimension primarily acts as an aggregation mechanism that consolidates token-level information into a compact patient-level representation, suppressing noise and redundancy without discarding critical features. Overall, these results demonstrate that the effectiveness of DimRed depends strongly on the axis and degree of compression, and that sequence-level aggregation provides a particularly robust representation for downstream classification.

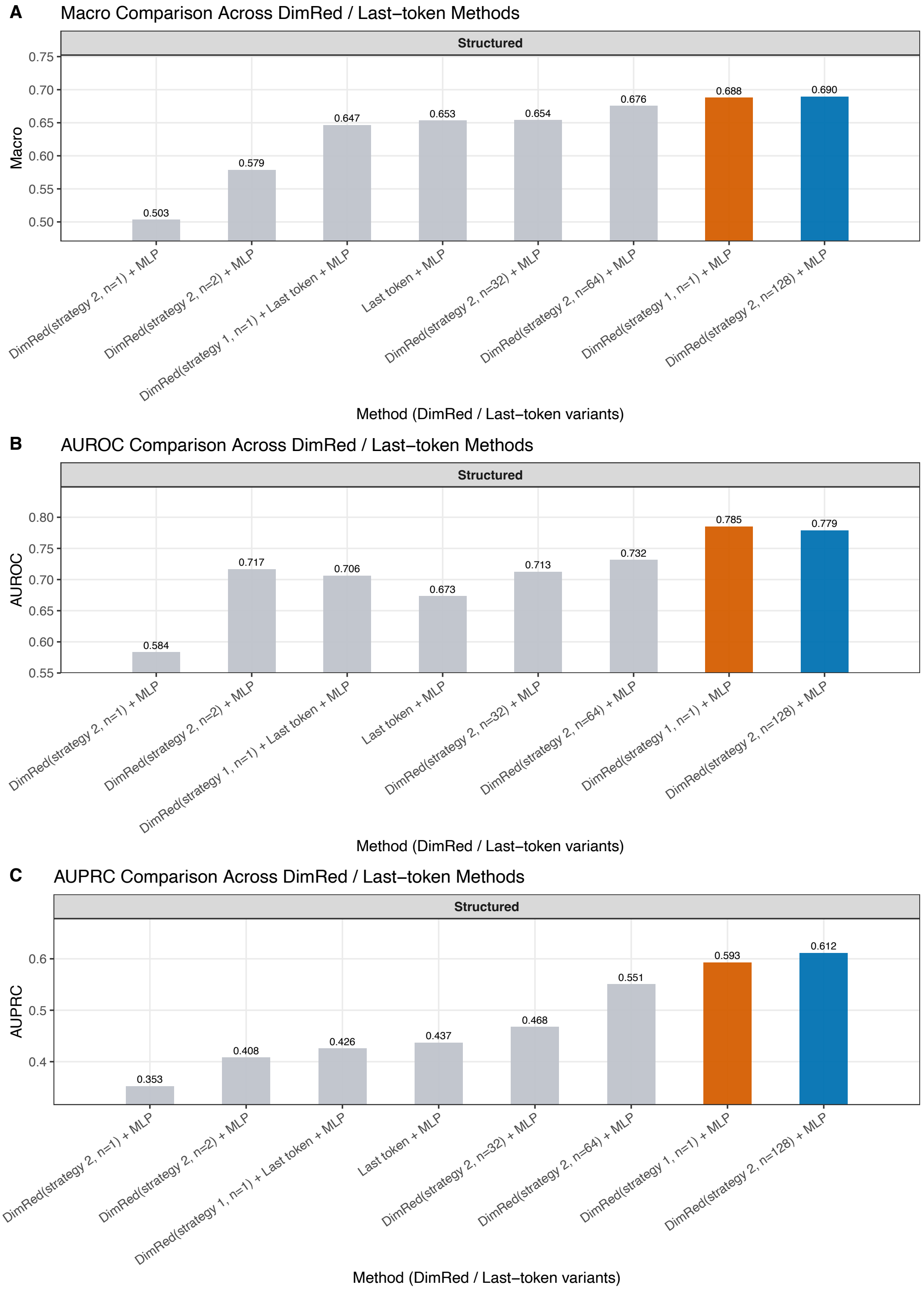

**Figure 3.** Task 3 results on MCPMD: Comparison of dimensionality reduction and last-token representation strategies within the RAG-MLP pipeline. This figure evaluates different representation strategies within a fixed RAG-MLP pipeline using structured EHR inputs. Panels (A–C) report performance measured by Macro-F1, AUROC, and AUPRC, respectively. Two dimensionality-reduction strategies are compared. Strategy 1 (DimRed on sequence length) applies dimensionality reduction across the token dimension to compress long sequences into a smaller number of token representations, emphasizing aggregation across clinical records. Strategy 2 (DimRed on hidden dimension) applies dimensionality reduction to the embedding dimension of token representations while preserving the full sequence length, focusing on feature-level compression. In addition, last-token representations are included as a baseline that summarizes each sequence using a single token embedding. Each bar corresponds to a specific representation variant, allowing direct comparison of how different compression strategies affect downstream patient–trial matching performance.

*4) Task 4: Frozen vs. Fine-tuned Representations*

We next compare frozen and fine-tuned LLM representations within identical RAG-based pipelines using mixed EHR inputs, where both structured records and clinical notes are jointly modeled (Figure 4). Across both evaluated pipeline variants (RAG-MLP and RAG-DimRed-MLP), fine-tuned representations consistently outperform frozen representations on Macro-F1, AUROC, and AUPRC. The performance gap is observed across all metrics, with particularly noticeable improvements in Macro-F1 and AUPRC, indicating enhanced sensitivity to the eligible class under class imbalance. While frozen representations provide reasonable baseline performance, their limitations become evident in the mixed setting, where effective integration of structured and unstructured information is required. In contrast, fine-tuning enables the model to adapt representations to domain-specific clinical language and task-relevant patterns, resulting in more robust and consistently improved downstream performance.

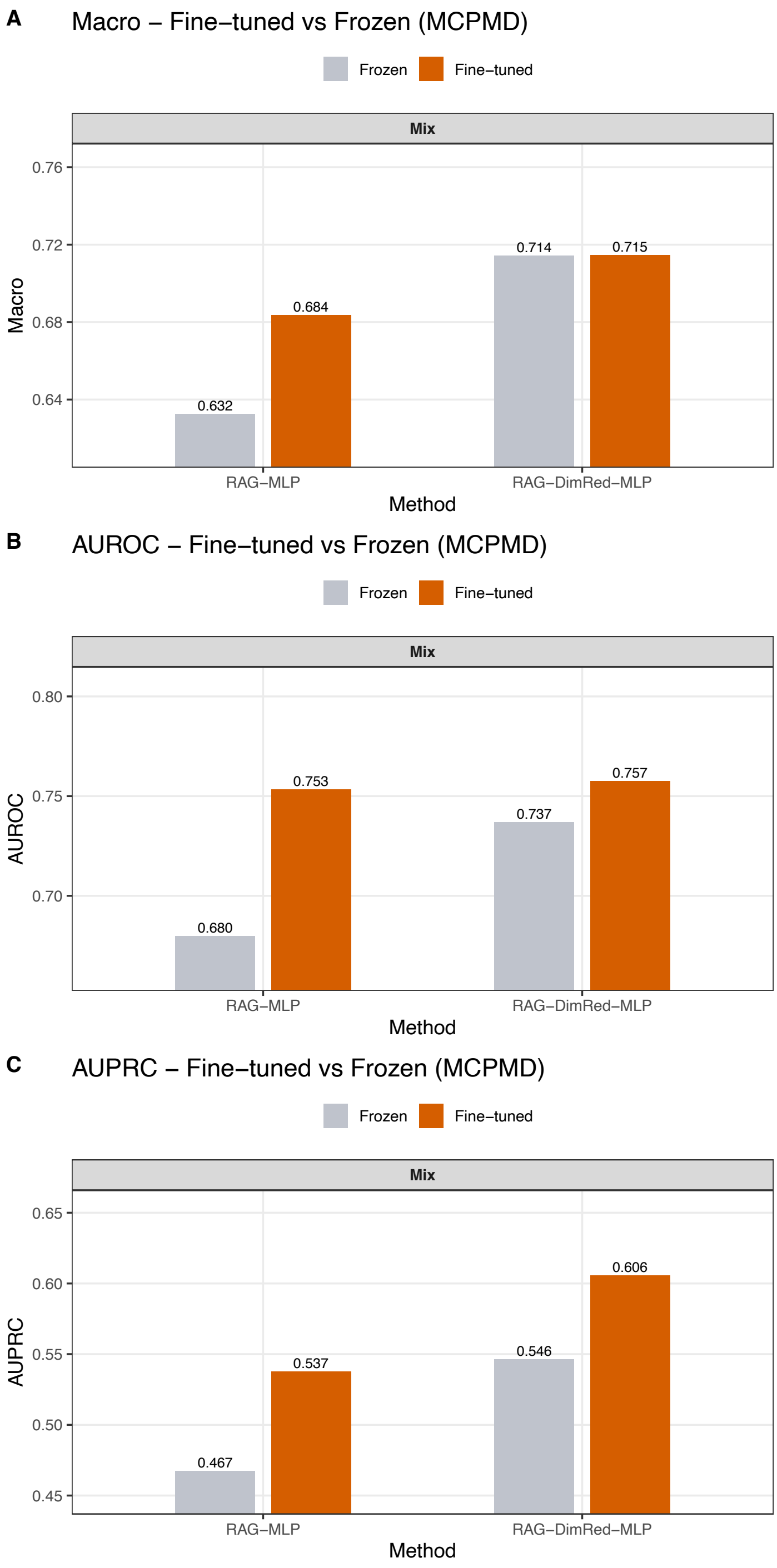

**Figure 4.** Task 4 results on MCPMD: Comparison of frozen and fine-tuned LLM representations within the RAG-based pipelines on mixed EHR data. This figure compares the performance of frozen and fine-tuned LLM representations within fixed RAG-based patient–trial matching pipelines on the MCPMD dataset using mixed EHR inputs. Panels (A–C) report results measured by Macro-F1, AUROC, and AUPRC, respectively. Two pipeline variants are evaluated, RAG-MLP and RAG-DimRed-MLP, with all components held constant except whether the LLM parameters are frozen or fine-tuned. This comparison isolates the effect of representation adaptation on downstream matching performance under otherwise identical pipeline settings.

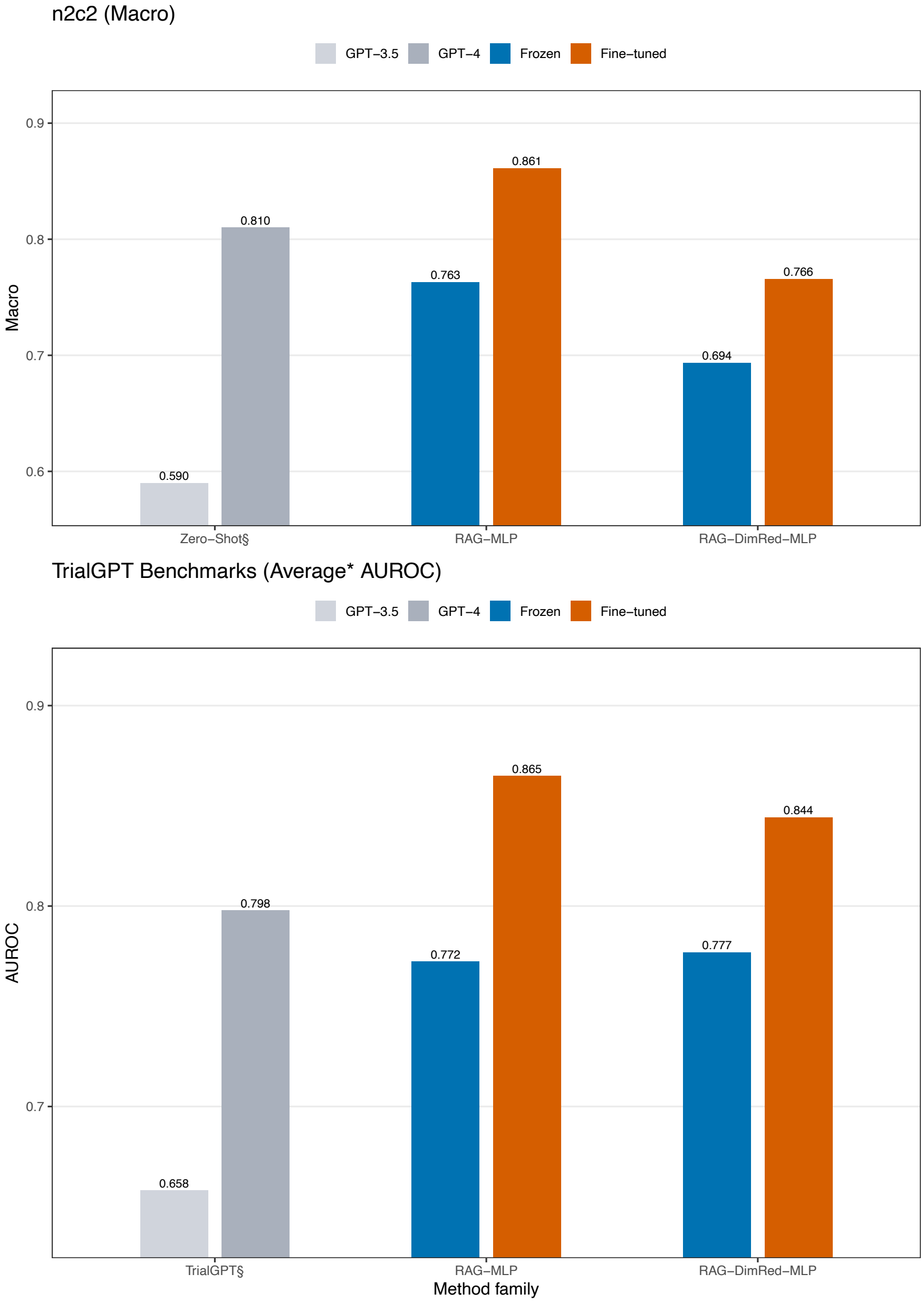

*SIGIR, TREC 2021, TREC 2022 (Average)

§Results were recorded from the original paper

**Figure 5.** Task 5 results: Comparison with zero-shot and TrialGPT patient–trial matching methods on open-source datasets. This figure compares the proposed RAG-based pipelines with zero-shot and TrialGPT patient–trial matching methods on four open-source benchmarks, including n2c2, SIGIR, TREC 2021, and TREC 2022 (average performance reported). The top panel reports Macro-F1 on the n2c2 dataset, while the bottom panel reports average AUROC across the SIGIR and TREC benchmark datasets. Results for zero-shot and TrialGPT methods are taken from the original publications, while results for RAG-MLP and RAG-DimRed-MLP are obtained from our experiments under both frozen and fine-tuned settings within an otherwise fixed pipeline. This comparison evaluates the performance of the proposed retrieval-augmented pipelines relative to existing TrialGPT and zero-shot baselines on open-source datasets.

### 5) *Task 5: Generalization Across Public Benchmarks*

We further evaluate the generalization ability of the proposed framework on four widely used open-source benchmarks, n2c2, SIGIR 2016, TREC 2021, and TREC 2022, which primarily consist of unstructured clinical notes. As shown in Figure 5, clear performance differences are observed across model variants and evaluation settings. On the n2c2 dataset, performance is reported using Macro-F1, while for SIGIR and TREC benchmarks, AUROC is used to align with prior studies. Across all datasets, pipelines using fine-tuned LLM representations consistently outperform their frozen counterparts, demonstrating improved modeling of clinical language and eligibility-related patterns. In addition, the proposed RAG-MLP and RAG-DimRed-MLP pipelines achieve competitive or superior performance compared with both zero-shot methods and TrialGPT, indicating that retrieval-augmented representations combined with lightweight classifiers provide an effective alternative to direct end-to-end matching approaches. The observed gains are particularly pronounced on datasets with annotation schemes and note structures similar to those in MCPMD, highlighting the importance of domain adaptation when transferring models to new clinical text collections.

### 6) *Task 6: Cross-Trial Generalization*

We conducted a series of cross-trial experiments to evaluate the model's ability to generalize to a target clinical trial when data from that trial were partially or fully excluded during training. Specifically, for each of the five selected trials from MCPMD (NCT01767909, NCT02008357, NCT02565511, NCT02669433, and NCT04468659), we progressively excluded 100% to 20% of that trial's samples from the training set and evaluated performance on the held-out trial. All experiments were performed using the mixed-data setting, which combines structured records and clinical notes. As shown in Figure 6, when a target trial is entirely excluded from training, model performance deteriorates substantially across all evaluation metrics. As the proportion of trial-specific data included during training increases, performance improves steadily, indicating that exposure to trial-specific data plays a critical role in downstream prediction. These results suggest that each trial represents a distinct data distribution with unique clinical characteristics, and insufficient representation of a trial during training leads to pronounced performance degradation when evaluated on that trial. Further analysis reveals that the three evaluation metrics respond differently to trial exclusion and domain shift. Macro-F1 shows the most pronounced decline when a trial is absent from training, reflecting its sensitivity to reduced recall on the minority (eligible) class under class imbalance. AUPRC exhibits a similarly strong drop, driven by degraded precision–recall behavior in the positive class. In contrast, AUROC degrades more gradually, as it measures global ranking performance and is less sensitive to threshold-dependent classification errors. Together, these findings highlight the importance of trial diversity in the training data and underscore the challenges of cross-trial generalization in real-world patient–trial matching.

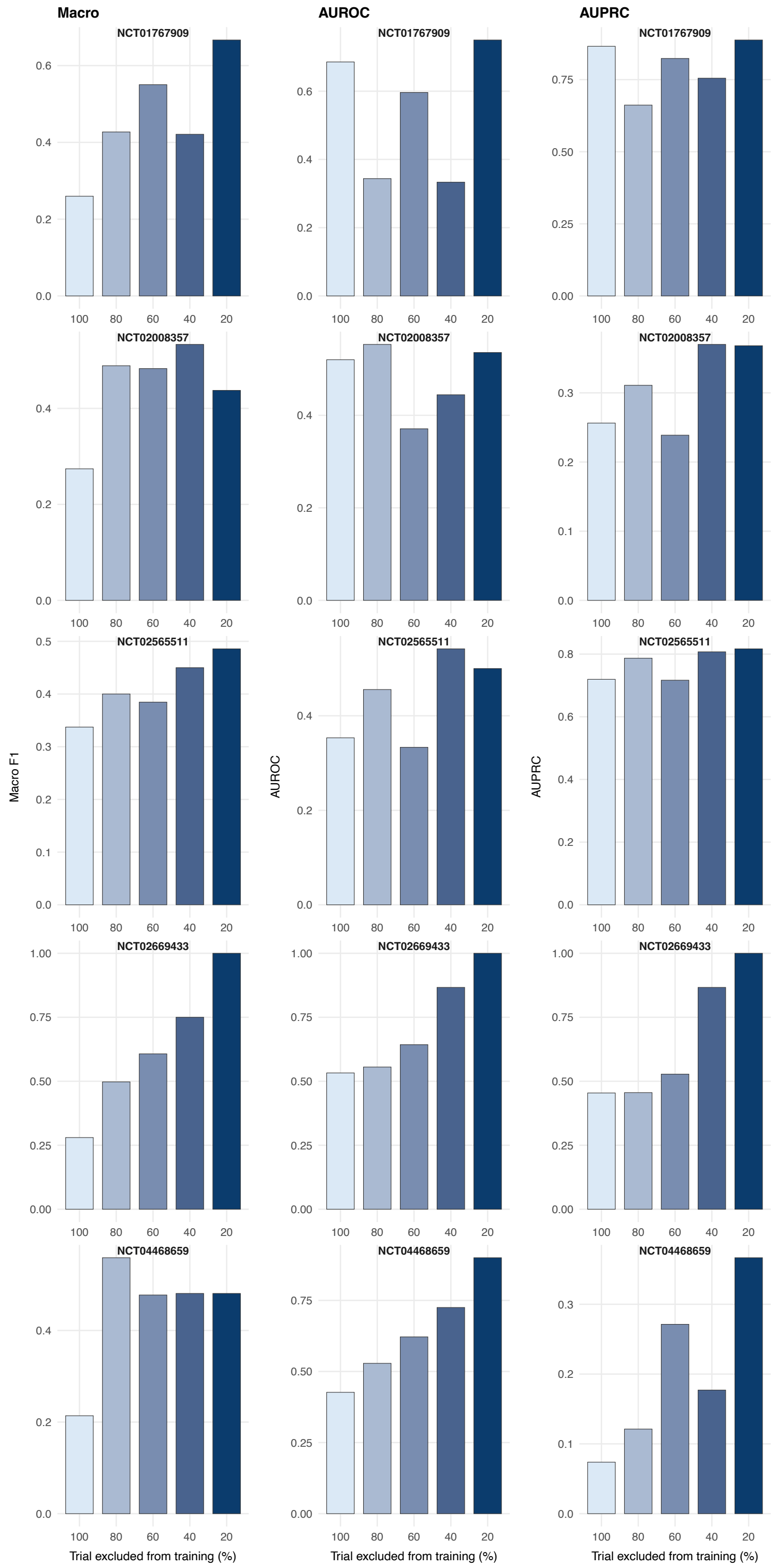

**Figure 6.** Task 6 results on MCPMD: Cross-trial evaluation of the RAG-based pipeline under varying levels of trial exclusion. This figure evaluates cross-trial generalization performance on the MCPMD dataset by systematically excluding varying proportions of data from a target trial during training. Each row corresponds to a held-out clinical trial, while columns report performance measured by Macro-F1, AUROC, and AUPRC, respectively. The x-axis denotes the percentage of samples from the target trial excluded from the training set, and the y-axis reports the corresponding performance metric on the held-out trial. Bars represent results obtained using the same RAG-based patient–trial matching pipeline, highlighting how model performance degrades or remains robust as trial-specific information is progressively removed from training. This experiment assesses the ability of the proposed approach to generalize across trials with limited or no trial-specific supervision.

### 7) Summary of Findings

Across all six tasks, we observe several consistent and complementary patterns that together characterize effective patient–trial matching with large language models. First, while pretrained LLMs provide reasonable representations for structured EHR data, they struggle to generalize to unstructured clinical text and mixed-modality inputs without task-specific adaptation. Fine-tuning substantially improves performance in these settings, particularly for metrics sensitive to class imbalance such as Macro-F1 and AUPRC, highlighting the importance of domain adaptation for clinical language. Second, dimensionality reduction emerges as a critical component of the modeling pipeline: appropriate compression and aggregation of high-dimensional representations stabilizes downstream learning, reduces noise, and improves performance across data modalities. Third, retrieval-augmented pipelines consistently outperform simpler baselines by enabling focused use of clinically relevant information without processing entire longitudinal records. Finally, cross-trial experiments demonstrate that generalization is highly sensitive to distributional differences across trials. Performance degrades sharply when trial-specific data are absent during training and improves steadily as exposure to each trial's data increases. This behavior underscores the heterogeneity of real-world clinical trials and suggests that robust patient–trial matching requires training data that adequately cover diverse trial designs, eligibility criteria, and patient populations. Taken together, these findings highlight the importance of combining retrieval augmentation, representation compression, and domain-specific adaptation to achieve scalable and generalizable patient–trial matching in real-world clinical settings.

## 3. Discussion

In this work, we proposed a patient–trial matching framework that integrates retrieval-augmented generation (RAG) with lightweight LLMs. By reducing input length through retrieval and leveraging locally deployed open-source LLMs, our method balances efficiency, privacy, and performance, providing a secure alternative to commercial black-box APIs. Extensive experiments on both multimodal real-world datasets (MCPMD) and widely used open-source benchmarks (n2c2, SIGIR, TREC 2021/2022) demonstrate consistent trends. Frozen LLMs provide useful inductive biases for structured records, but fine-tuning is indispensable for unstructured clinical notes, leading to SOTA performance. At the same time, cross trials evaluations highlight the challenges of generalization across heterogeneous protocols. Our study demonstrates the effectiveness of integrating RAG with lightweight LLMs for the task of clinical trial patient matching. Across multiple datasets and modalities, several consistent findings emerge.

Compared to traditional machine learning methods (e.g., random forest, decision tree, and SVM) and early NLP-based systems such as Criteria2Query and EliIE, our framework demonstrates greater robustness and adaptability. Unlike commercial-model approaches such as TrialGPT, our locally deployed pipeline reduces the risk of data leakage and ensures greater control over data storage and model updates, thereby addressing critical privacy and security concerns in healthcare applications. At the same time, by leveraging RAG to reduce input length and computational costs, we show that lightweight open-source LLMs can deliver competitive performance without relying on prohibitively expensive trillion-parameter models or commercial APIs.

For the performance of models, first, frozen LLMs exhibit useful inductive biases when applied to structured EHR data, and techniques such as DimRed-enhanced CLS embeddings further boost their performance. By replacing different LLMs, it is basically proved that the prior knowledge of LLM is capable of extracting effective embeddings from medical data. However, frozen models fail to generalize effectively to unstructured clinical notes, where fine-tuning becomes indispensable. This conclusion is acceptable because most of the current LLM work requires fine-tuning models on specific downstream tasks. Second, when structured and unstructured components are combined, the performance patterns are dominated by the unstructured modality, highlighting the central role of free-text narratives in patient–trial matching. Finally, results on open-source datasets such as n2c2, SIGIR 2016, and TREC 2021/2022 further confirm that fine-tuning is critical for achieving SOTA performance, as pretrained knowledge alone does not transfer effectively to clinical notes.

Despite the improvements observed across multiple tasks, several limitations remain. First, our evaluation focuses on trial-level labels and matching outcomes rather than real-world enrollment decisions, which are influenced by additional operational constraints (e.g., site capacity, clinician preference, and evolving protocol interpretations). As a result, strong predictive performance does not necessarily imply direct clinical utility without prospective validation.

Second, the learning objective and metrics are sensitive to label definitions and class imbalance. Open-source benchmarks use heterogeneous label taxonomies (e.g., “potential/eligible” vs. “excluded/ineligible”), and converting them into a binary task may discard clinically meaningful distinctions. Moreover, Macro-F1 and AUPRC emphasize minority-class behavior; while appropriate for matching, they can fluctuate substantially under small changes in positive prevalence, making cross-dataset and cross-trial comparisons less stable.

Third, the framework relies on a fixed retrieval-and-representation pipeline whose behavior may vary across trials and institutions. In MCPMD, trials differ in protocol structure and documentation practices, and our cross-trial experiments show performance drops when a trial is unseen during training. This suggests that distribution shift across trials remains a major challenge and that robustness depends on sufficient coverage of trial-specific language and patterns during training.

Fourth, our method compresses high-dimensional representations (via DimRed and classifier heads), which introduces design dependencies that may not transfer universally. The best-performing DimRed configurations in Task 3 may be specific to the structured setting and the current backbone/pipeline choices; different note lengths, tokenization behaviors, or trial compositions could change the optimal compression strategy. More systematic analysis of hyperparameter sensitivity and calibration across settings is needed.

Future work will focus on improving robustness under trial and dataset shift (e.g., stronger cross-trial adaptation, better handling of heterogeneous label schemas, and evaluation under more realistic deployment scenarios), as well as extending the framework to additional modalities and broader patient populations to better reflect real-world trial screening workflows.

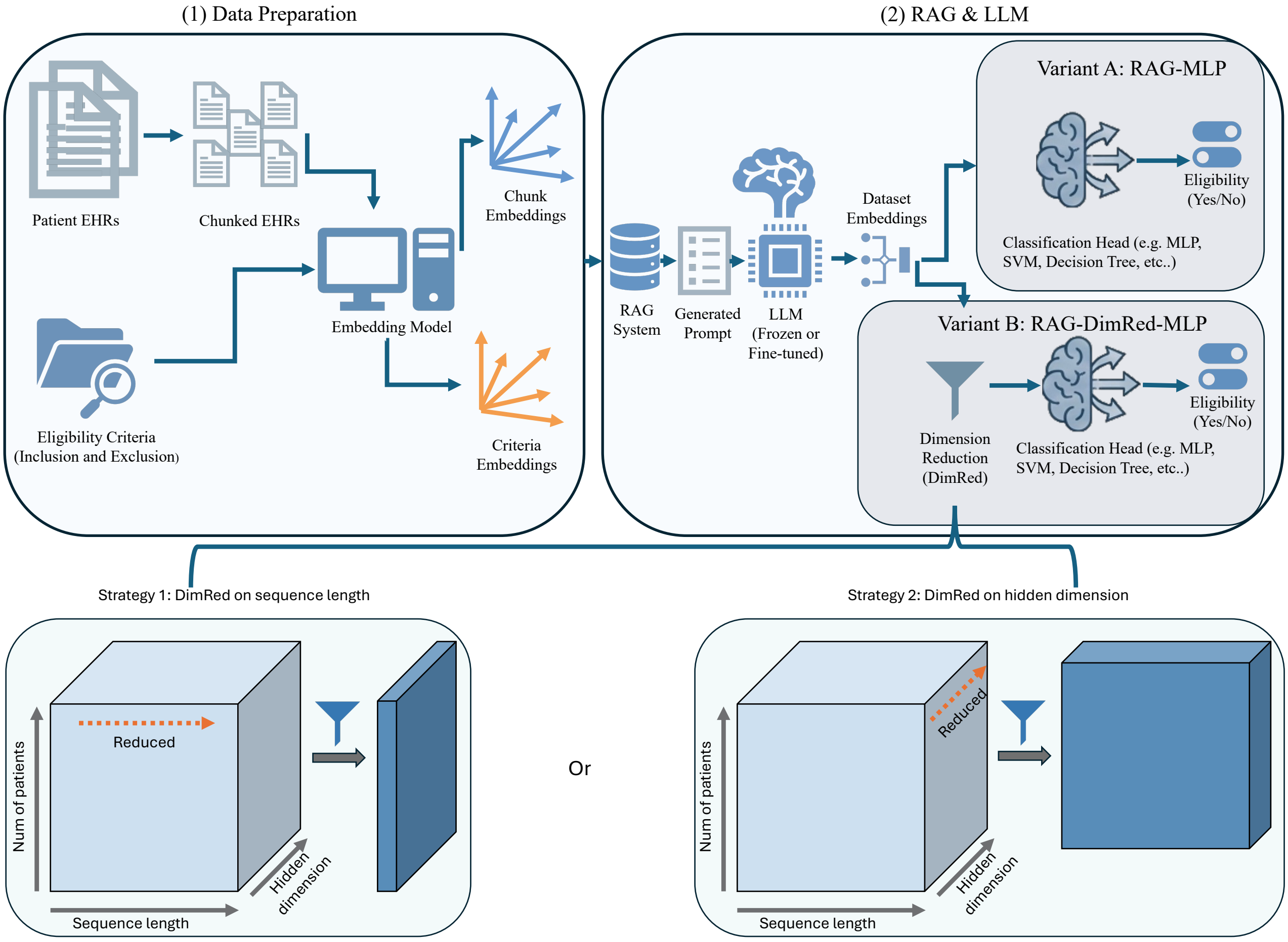

**Figure 7.** Overview of the Lightweight RAG-LLM Pipeline for Patient–Trial Matching. The framework illustrates a lightweight two-stage RAG + LLM pipeline for scalable patient–trial eligibility classification, followed by two efficient downstream model variants. In Stage (1) Data Preparation, patient EHRs are split into textual chunks, and both the chunked EHRs and clinical trial eligibility criteria (inclusion and exclusion statements) are encoded to generate compact embeddings. In Stage (2) RAG & LLM, these embeddings are stored in a retrieval system, which selects clinically relevant context instead of processing full records, enabling efficient input construction. The retrieved information is assembled into prompts and passed to a frozen or fine-tuned LLM to produce lightweight dataset embeddings. These representations are then processed by two efficient classifier variants: Variant A (RAG-MLP), where embeddings are directly input into a lightweight classification head (e.g., MLP, SVM, or decision tree), and Variant B (RAG-DimRed-MLP), where a dimensionality reduction (DimRed) module further compresses embeddings before classification, improving efficiency while preserving predictive signal. Two DimRed strategies are considered: reduction along the sequence-length dimension and reduction along the hidden-dimension axis. Overall, the framework achieves lightweight modeling through retrieval-based information selection, compact representation learning, and efficient downstream prediction, enabling scalable and practical deployment.

## 4. Methods

Figure 7 presents an overview of the lightweight RAG–LLM framework for patient–trial matching. Patient EHRs are split into textual chunks and encoded together with trial eligibility criteria to generate chunk and criteria embeddings used by the RAG module, enabling efficient selection of clinically relevant information instead of processing full records. In the first configuration, prompts derived from the retrieved data are fed into a frozen or fine-tuned LLM, whose high-dimensional embeddings are passed directly to a lightweight classification head such as an MLP, SVM, or decision tree. In the second configuration, the LLM still generates embeddings, but a DimRed module is introduced to compress representations, reducing dimensionality before classification and improving computational efficiency while preserving essential information. Both configurations share the same upstream data flow and differ only in how LLM embeddings are processed for downstream prediction. We consider two dimensionality-reduction strategies: one that operates along the sequence-length dimension and another that operates along the hidden-state dimension, both contributing to compact and efficient representation learning.

### 4.1. RAG modules

For the patient matching dataset, we have a set of patients $P$ and their corresponding EHRs $N$, where each patient $p \in P$ has a set of note $n_p \in N$. The target is to get the top-k most relevant EHRs. The eligibility criteria contain both inclusion and exclusion conditions, denoted as $C$ = {c1, c2, …, cm}. Based on an embedding model(For example, BioBERT[19] for our paper) M, we can map the text t from EHRs and eligibility criteria into a vector space, shown as M: $t \rightarrow R^d$, where the $d$ refers to the dimensionality of the embedding space. We can get the embeddings vn and vc from EHRs and eligibility criteria, where vn = M($n_p$) and vc = M(cm). To rank the relevance of patient notes to the trialcriteria, we compute the cosine similarity between each note embedding and the criterion embedding: $\langle v_n, v_c \rangle = \frac{v_n \cdot v_c}{\|v_n\| \|v_c\|}$. The cosine similarity score can be ranked to get the top-k most relevant notes. In our experiments, we set k=4 based on empirical validation.

### 4.2. Prompt Template for LLM

After get the most relevant chunks of the EHRs: $\left\{n_p^{(k)}\right\} = argmax \sum_{c \in C} \langle v_n, v_c \rangle$. We combine them with the trial eligibility criteria and system instructions to construct a structured input prompt I. The system instructions provide guidance on how the LLM should understand the patient records based on the given eligibility criteria.

### 4.3. The Embedding of Chunked Prompt from LLM

The LLM (We use the mistral 7B, Llama 3 8B, and Falcon 7B) h receives these structured prompts, which are the most relevant EHRs chunk without the unnecessary information, reducing the computational overhead, then output the embedding (or the hidden state) based on its prior knowledge.

We can get the chunk embedding: $v_I = h(I) \in R^{l \times k}$, where l refers to the number of the tokens, and the k is the dimension of the last hiddent layer of the LLM h.

The directly obtained chunk embeddings have a shape of l×k, where l represents the number of tokens. Since not all tokens can adequately represent the information contained in the EHR chunk, we consider three processing strategies: 1) We aggregate token embeddings by averaging along the sequence dimension: $\overline{v_I} = \frac{\sum_i^l v_{I,i}}{l}$. 2) We first apply dimensionality reduction to capture the dominant variation in the token embeddings and then aggregate the reduced representations. Specifically, the token embeddings are mean-centered: $\widetilde{v_I} = v_I - \overline{v_I}$, and the covariance matrix is computed as $T = \frac{1}{l-1} \widetilde{v_I}^{\top} \widetilde{v_I}$ . We perform eigen-decomposition $T v_{I,i} = \lambda_i v_{I,i}$ and select the top-k principal components based on the variance (eigenvalues) to form the projection matrix W. The centered embeddings are projected into the reduced space, $Z = \widetilde{v_I} \mathrm{W}, \mathrm{Z} \in \mathrm{R}^{l \times k}$ and average it to $\mathrm{R}^k$. 3) We use the embedding of the special CLS token embedding from the LLM as a compact representation of the entire chunk. Finally, we can reduce the dimension of chunk embedding from an l×k to k.

### 4.4. Dimension Reduction (DimRed) Module

After generating embeddings with the LLM, we apply DimRed either along the sequence-length dimension or along the transposed hidden dimension with a specified number of components. This reduces the embedding dimensionality while preserving the key information contained in the dataset.

### 4.5. Classifier

After obtaining the chunk embeddings, we define a fully connected layer f as the classifier. Based on the ground-truth labels $y_p$, the training process of the classifier is to optimize the model parameters θ by minimizing the binary cross-entropy loss function:$L(\theta) = -\sum_{p \in P} \left[ y_p \log f_\theta(v_I) + \left(1 - y_p\right) \log(1 - f_\theta(v_I)) \right]$. We employ gradient-based optimization Adam optimizer[20] for updating the parameters of the classifier, denoted as: $\theta^{(t+1)} = \theta^{(t)} -$ $\eta \nabla L(\theta^{(t)})$, where $\eta$ is the learning rate, $\nabla$ means the gradient of the loss function. The training process last on training set until convergence, ensuring the model generalizes well to unseen patients (test set).

### 4.6. Workflow

We designed six sets of experiments, corresponding to Tasks 1–6, to systematically evaluate the effects of downstream modeling choices, LLM backbone selection, fine-tuning, dimensionality reduction, cross-dataset generalization, and cross-trial robustness within a unified RAG-based patient–trial matching framework. Across all experiments, EHRs and eligibility criteria were first processed by the same RAG module to construct chunked prompts, which were encoded into LLM representations. Differences between experiments arise only from how these representations are processed, trained, or evaluated downstream.

1) *Effect of Downstream Classification Strategy (Task 1)*

To evaluate the impact of downstream classifiers, we applied different classification methods to the same frozen LLM representations. RAG-encoded inputs were passed through a frozen LLM, after which the LLM output layer was removed and replaced by either traditional machine learning classifiers, Random Forest (RF), Decision Tree (DT), and Support Vector Machine (SVM), or a lightweight MLP classifier. This experiment isolates the effect of downstream classification strategy while holding representation construction fixed.

2) *Effect of LLM Backbone Choice (Task 2)*

To assess the robustness of the RAG-based representations to the choice of LLM backbone, we evaluated multiple frozen LLMs, including Mistral-7B, Falcon-7B, and Llama3-8B, within an otherwise fixed pipeline. RAG-encoded inputs were fed into each frozen LLM, and the resulting embeddings were passed to the same MLP classifier. All other components were held constant.

3) *Effect of Dimensionality Reduction Strategy (Task 3)*

To study the impact of representation compression, we evaluated different dimensionality reduction (DimRed) strategies applied to frozen LLM embeddings. These included DimRed along the sequence-length dimension, DimRed along the hidden-state dimension with varying numbers of components, last-token embeddings, and hybrid combinations. The compressed embeddings were then fed into a fixed downstream classifier. This experiment examines how different compression axes and degrees affect downstream performance.

*4) Effect of Fine-tuning vs. Frozen (Task 4)*

To evaluate the benefit of representation adaptation, we compared frozen and fine-tuned LLM representations within identical RAG-based pipelines. In the frozen setting, only the downstream MLP classifier was trained, while in the fine-tuned setting, the LLM and the MLP classifier were updated jointly. Both RAG-MLP and RAG-DimRed-MLP pipelines were evaluated to isolate the effect of fine-tuning under consistent pipeline structures.

*5) Generalization Across Public Benchmarks (Task 5)*

To assess generalization beyond MCPMD, we evaluated the proposed pipelines on four open-source datasets, n2c2, SIGIR 2016, TREC 2021, and TREC 2022, which primarily consist of unstructured clinical notes. Performance was compared against zero-shot methods and TrialGPT using the evaluation metrics reported in the original studies. This experiment examines cross-dataset generalization under consistent pipeline settings.

*6) Cross-Trial Generalization (Task 6)*

Finally, to evaluate robustness across clinical trials, we conducted cross-trial experiments on MCPMD by progressively excluding 100% to 20% of samples from a target trial during training and evaluating performance on that trial. All experiments were performed using the mixed-data setting. This experiment assesses the sensitivity of the model to trial-specific distribution shift and the extent to which exposure to trial-specific data is required for reliable performance.

## Acknowledgments

This work is supported by a grant from the National Institute of Health (NIH) NIGMS (R00GM135488).

## Data Availability

The open-source datasets analyzed in this study (n2c2, SIGIR, TREC 2021/2022) are publicly available from their respective shared-task repositories under standard data-use agreements. The Mayo Clinic Platform data used in this study are not publicly available due to patient privacy and institutional restrictions. Derived data supporting the findings are available from the corresponding author upon reasonable request and subject to data-governance requirements.

## Author Contributions

N.Z. conceptualized and supervised the whole study. N.Z. led and managed this project. X.L. and Y.X. contributed to data processing, patient matching task design, and experiment. M.L. contributed to the optimization of the experiments. N.Z., X.L., and Y.X. composed the initial draft of the report, and all authors reviewed the manuscript. The final manuscript was approved by all authors.

## References

1. Wornow M, Lozano A, Dash D, Jindal J, Mahaffey KW, Shah NH. Zero-shot clinical trial patient matching with llms. NEJM AI. 2025;2(1):AIcs2400360.

2. Jin Q, Wang Z, Floudas CS, Chen F, Gong C, Bracken-Clarke D, et al. Matching patients to clinical trials with large language models. Nature communications. 2024;15(1):9074.

3. Li X, Chowdhury S, Wi CI, Vassilaki M, Liu X, Sio TT, et al. LLM-Match: An Open-Sourced Patient Matching Model Based on Large Language Models and Retrieval-Augmented Generation. arXiv preprint arXiv:250313281. 2025.

4. Hutson M. How AI is being used to accelerate clinical trials. Nature. 2024;627(8003):S2-S5.

5. Woo M. An AI boost for clinical trials. Nature. 2019;573(7775):S100-S.

6. Shivade C, Raghavan P, Fosler-Lussier E, Embi PJ, Elhadad N, Johnson SB, et al. A review of approaches to identifying patient phenotype cohorts using electronic health records. Journal of the American Medical Informatics Association. 2014;21(2):221-30.

7. Food, Administration D. Enhancing the diversity of clinical trial populations—eligibility criteria, enrollment practices, and trial designs guidance for industry. 2020.

8. Xu S, Yang L, Kelly C, Sieniek M, Kohlberger T, Ma M, et al. Elixr: Towards a general purpose x-ray artificial intelligence system through alignment of large language models and radiology vision encoders. arXiv preprint arXiv:230801317. 2023.

9. Kang T, Zhang S, Tang Y, Hruby GW, Rusanov A, Elhadad N, et al. EliIE: An open-source information extraction system for clinical trial eligibility criteria. Journal of the American Medical Informatics Association. 2017;24(6):1062-71.

10. Yuan C, Ryan PB, Ta C, Guo Y, Li Z, Hardin J, et al. Criteria2Query: a natural language interface to clinical databases for cohort definition. Journal of the American Medical Informatics Association. 2019;26(4):294-305.

11. Gehrmann S, Dernoncourt F, Li Y, Carlson ET, Wu JT, Welt J, et al. Comparing deep learning and concept extraction based methods for patient phenotyping from clinical narratives. PloS one. 2018;13(2):e0192360.

12. Leaman R, Khare R, Lu Z. Challenges in clinical natural language processing for automated disorder normalization. Journal of biomedical informatics. 2015;57:28-37.

13. Achiam J, Adler S, Agarwal S, Ahmad L, Akkaya I, Aleman FL, et al. Gpt-4 technical report. arXiv preprint arXiv:230308774. 2023.

14. Touvron H, Lavril T, Izacard G, Martinet X, Lachaux M-A, Lacroix T, et al. Llama: Open and efficient foundation language models. arXiv preprint arXiv:230213971. 2023.

15. Stubbs A, Filannino M, Soysal E, Henry S, Uzuner Ö. Cohort selection for clinical trials: n2c2 2018 shared task track 1. Journal of the American Medical Informatics Association. 2019;26(11):1163-71.

16. Koopman B, Zuccon G, editors. A test collection for matching patients to clinical trials. Proceedings of the 39th International ACM SIGIR conference on Research and Development in Information Retrieval; 2016.

17. Roberts K, Demner-Fushman D, Voorhees EM, Bedrick S, Hersh WR, editors. Overview of the TREC 2021 Clinical Trials Track: TREC; 2021.

18. Roberts K, Demner-Fushman D, Voorhees EM, Bedrick S, Hersh WR, editors. Overview of the TREC 2022 Clinical Trials Track: TREC; 2022.

19. Lee J, Yoon W, Kim S, Kim D, Kim S, So CH, et al. BioBERT: a pre-trained biomedical language representation model for biomedical text mining. Bioinformatics. 2020;36(4):1234-40.

20. Kingma DP, Ba J. Adam: A method for stochastic optimization. arXiv preprint arXiv:14126980. 2014.